\documentclass[a4paper,conference]{IEEEtran}
\usepackage{cite}

\ifCLASSINFOpdf
  \usepackage[pdftex]{graphicx}
  \graphicspath{{./img/}}
  \DeclareGraphicsExtensions{.pdf,.jpeg,.png}
\else
  \usepackage[dvips]{graphicx}
  \graphicspath{{./img/}}
  \DeclareGraphicsExtensions{.eps}
\fi
\usepackage{amssymb,amsfonts}
\usepackage{amsmath}
\def \x {\mathbf{x}}

\def \w {\mathbf{w}}
\def \R {\mathbb{R}}
\def \E {\mathbb{E}}

\ifCLASSOPTIONcompsoc
 \usepackage[caption=false,font=normalsize,labelfont=sf,textfont=sf]{subfig}
\else
 \usepackage[caption=false,font=footnotesize]{subfig}
\fi

\usepackage{stfloats}
\usepackage{url}

\usepackage{multirow}

\begin{document}
%
\title{Few-Shot Font Generation \\ with Deep Metric Learning}




%
\author{\IEEEauthorblockN{Haruka Aoki,
Koki Tsubota,
Hikaru Ikuta and
Kiyoharu Aizawa}
\IEEEauthorblockA{Department of Information and Communication Engineering,\\
The University of Tokyo
\\ Email: \{h\_aoki, tsubota, ikuta, aizawa\}@hal.t.u-tokyo.ac.jp}
}


\maketitle

\begin{abstract}

  Designing fonts for languages with a large number of characters, such as Japanese and Chinese, is an extremely labor-intensive and time-consuming task.
  In this study, we addressed the problem of automatically generating Japanese typographic fonts from only a few font samples, where the synthesized glyphs are expected to have coherent characteristics, such as skeletons, contours, and serifs.
  Existing methods often fail to generate fine glyph images when the number of style reference glyphs is extremely limited.
  Herein, we proposed a simple but powerful framework for extracting better style features. This framework introduces deep metric learning to style encoders.
  We performed experiments using black-and-white and shape-distinctive font datasets and demonstrated the effectiveness of the proposed framework.

\end{abstract}


%
\IEEEpeerreviewmaketitle

\section{Introduction}
Desktop publishing has been widely used in the printing and publishing industry.
Moreover, numerous digital fonts are available for use for various purposes.
However, typographic fonts containing Japanese and Chinese characters are fewer in comparison with those containing Latin ones.
Consequently, the range of publication or advertisement designs in such languages is narrow. 
One of the reasons for the paucity of font diversity in East Asian languages is that designing typographic fonts is a highly labor-intensive and time-consuming task.
It requires the handicraft of a professional designer because of the complexity of the glyph shapes and the large number of characters used.
Approximately thousands of characters are required for daily use, and tens of thousands of characters are required to cover the entire language, in contrast to the 26 letters required for the Latin alphabet.

Therefore, automatic generation of glyph images from only a few references is required, especially for languages having a large number of characters.
Such automatic generation would reduce the workload of font designers and enable them to create more diverse and unique fonts.
It may also help non-professional users in building their original font library.

\begin{figure}[!t]
  \centering
  \includegraphics[width=3.3in]{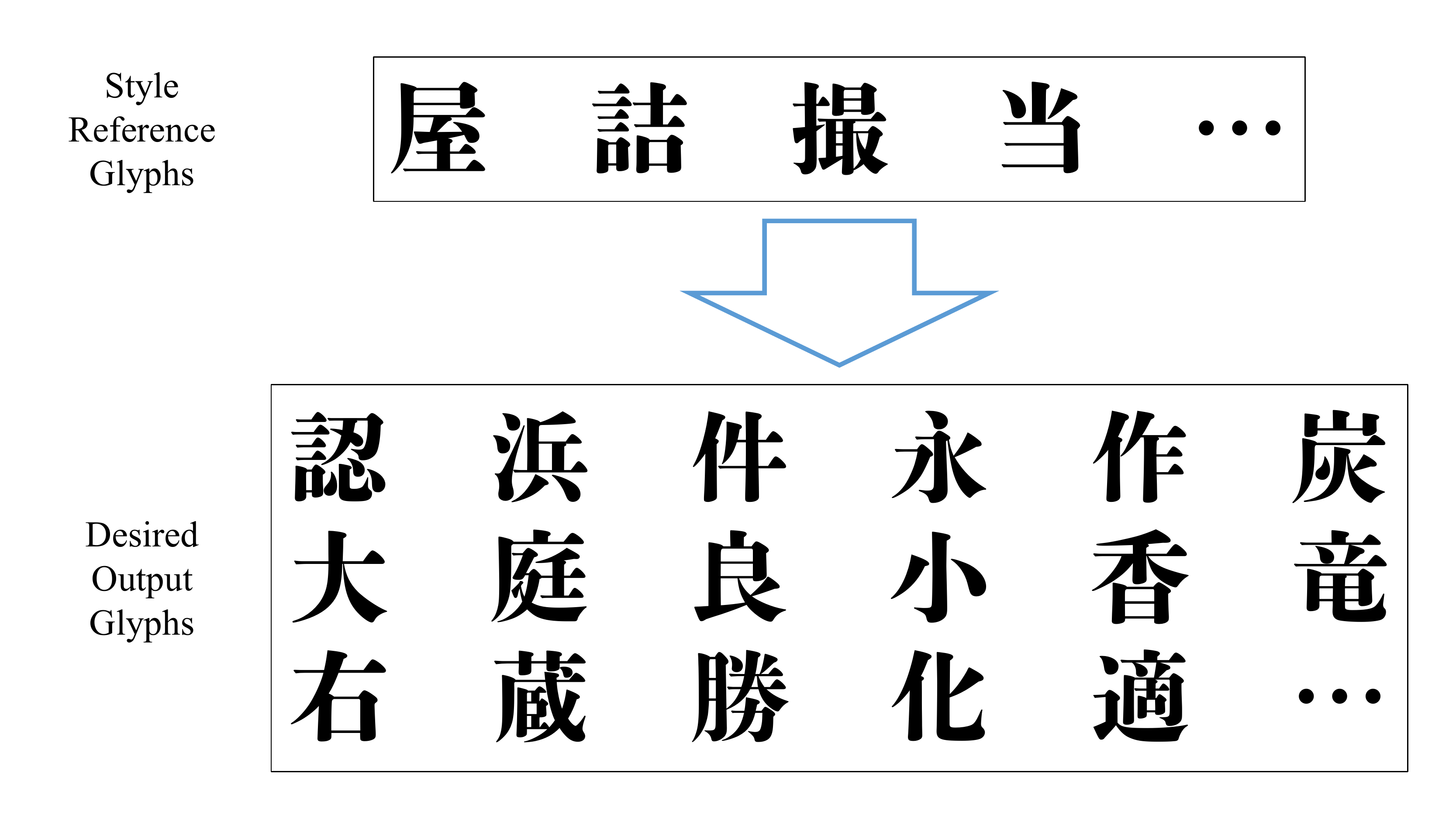}
  \caption{Overview of our few-shot font generation task}
  \label{fig:overview}
\end{figure}

This study addresses the problem of generating Japanese typographic fonts, which have a large number of characters, from only a few style reference glyphs as input (Fig. \ref{fig:overview}).
The generated glyphs are expected to have coherent styles with reference glyphs in terms of the shape of skeletons, contour of serifs, and thickness of lines.

Various attempts have been made thus far to design fonts easily.
Previous studies on font generation focused on the modeling of outlines \cite{suveeranont2010example,Campbell2014Learning,Zhou2011Easy}.
These earlier methods often required human intervention, and the obtained results were insufficient or not coherent among the characters constituting a single font set.

With the rise of deep learning, several end-to-end font generation methods have been proposed over the past few years \cite{upchurch2016z, zi2zi, Lyu2017Auto, jiang2017dcfont, jiang2019scfont}.
However, these methods require a large number of style reference glyphs (approximately hundreds), which may be arduous to obtain in some cases.

Recently, some studies have been conducted to generate fonts from a few reference glyphs without requiring a large number of references.
These studies can be classified into two types based on their trends: texture style transfer \cite{azadi2018multi, Gao2019Artistic, yang2019tet} and glyph shape transfer \cite{zhang2018separating, Sun2018Learning, cha2020few}.
The latter is generally a more challenging task than the former as glyph shape features (such as outlines and skeletons) are not easy to separate into content and style.
Although these methods successfully generate unseen style fonts with only a few reference glyphs, their results have room for improvement, especially when the number of style reference glyphs is extremely limited.

\begin{figure}[!t]
  \centering
  \includegraphics[width=3.3in]{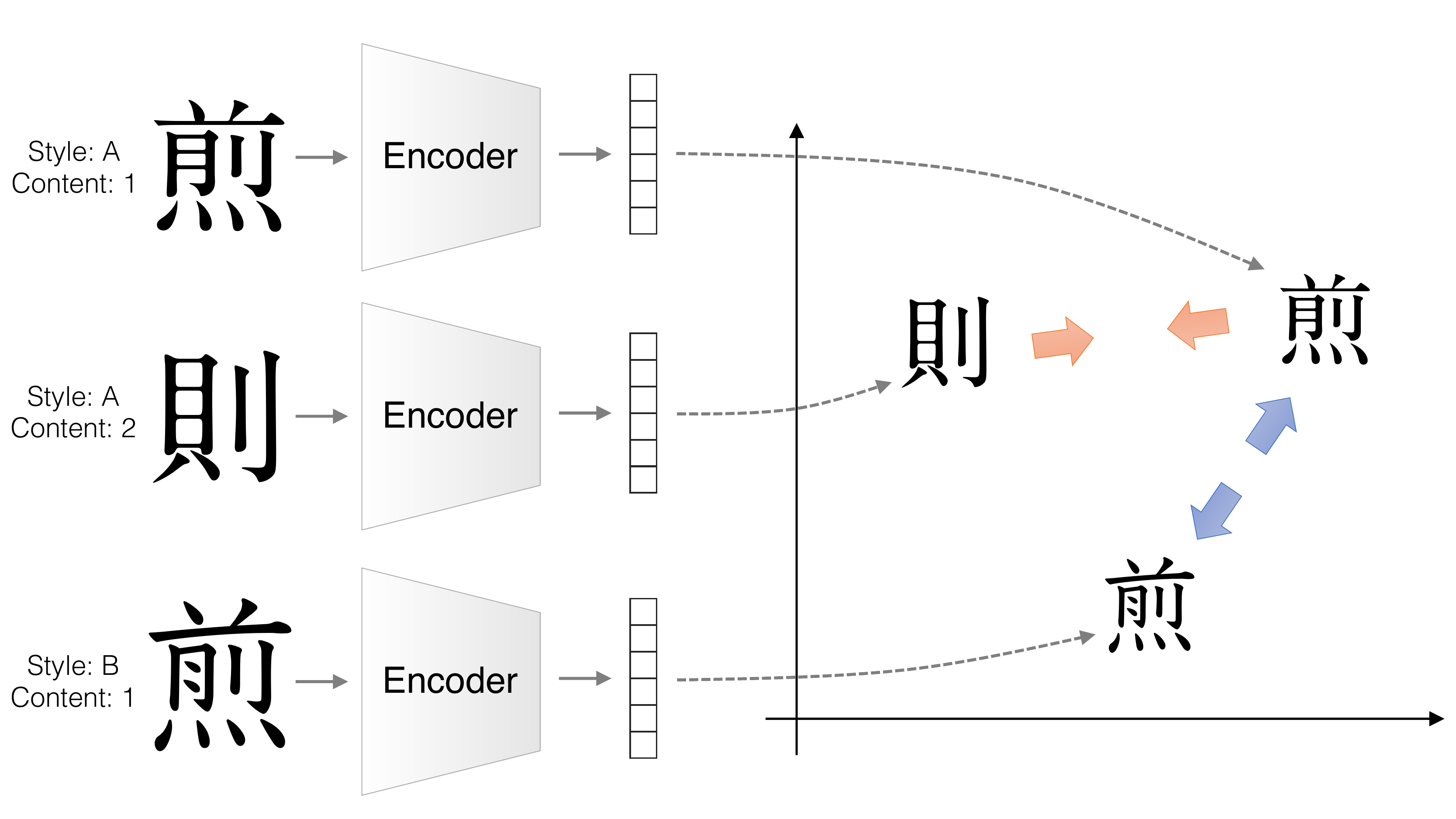}
  \caption{Extraction of only style features by the style feature encoder through metric learning. Glyphs with the same style should be closer to each other in the style feature space, whereas glyphs with different styles should be far away from each other (even when they have the same content).}
  \label{fig:metric}
\end{figure}

In this study, we focus on the common structure used in recent approaches \cite{zhang2018separating,Gao2019Artistic,yang2019tet,Sun2018Learning,cha2020few}: content encoder + style encoder + decoder(s). 
We propose a simple but effective framework to improve the performance of the existing font generation networks: applying metric learning to their style feature encoders.
This method forces the outputs of the style encoder to be closer to each other in the feature space when the input glyphs have the same style; otherwise, the outputs are far away from each other (Fig. \ref{fig:metric}).
Thus, this method allows the style feature encoders to extract only the style features while reducing the impact of the differences in the content features.

We select AGIS-Net \cite{Gao2019Artistic} and EMD \cite{zhang2018separating} as our baselines and backbone networks.
AGIS-Net is currently one of the best-performing models in the field of few-shot font generation.
However, it has the drawback of poor quality in generating black-and-white and shape-distinctive fonts.
EMD is an outstanding model for generating binary and shape-distinctive fonts.
However, it often fails to extract novel style features and produces collapsed glyphs.
We improve them by utilizing deep metric learning (DML) and discuss the effectiveness of our proposed framework.
In summary, our contributions are as follows:
\begin{enumerate}
  \item We introduce a simple DML method to existing font generation systems.
  \item We show that metric learning contributes to extracting better style feature embeddings and producing more prominent results, especially when the number of style reference glyphs is considerably limited.
\end{enumerate}

\section{Related Works}

\subsection{Font Generation}
Previous studies on automatic font generation focused on the explicit modeling of glyph outlines.
Suveeranont and Igarashi \cite{suveeranont2010example} automatically extracted a skeleton from the outline of a reference glyph and generated a consistent font set utilizing a weighted blend of template fonts.
Campbell and Kautz \cite{Campbell2014Learning} built a manifold of standard Latin fonts, on which we can generate fonts by interpolating existing fonts.
Zhou et al. \cite{Zhou2011Easy} designed glyph modeling for Chinese characters by utilizing radicals.
Lian et al. \cite{Lian2016Automatic} proposed a model to generate Chinese handwriting fonts with neural networks to learn styles.

With the rise of deep learning, end-to-end font generation methods have been proposed recently.
Upchurch et al. \cite{upchurch2016z} proposed a method to synthesize Latin fonts in a supervised way with variational autoencoders (VAEs) \cite{kingma2013auto}.
The zi2zi \cite{zi2zi} model utilizes generative adversarial networks (GANs) \cite{goodfellow2014generative} to generate Chinese typographic fonts.
Lyu et al. \cite{Lyu2017Auto} also used GANs to generate Chinese calligraphic images.
Jiang et al. proposed a system to synthesize Chinese handwriting fonts from 775 samples \cite{jiang2017dcfont} and later introduced a more powerful solution \cite{jiang2019scfont}.
However, these methods require hundreds or more style reference glyphs, which may be arduous to obtain in some cases, such as for handwriting fonts.

Several studies have been recently conducted to synthesize a font library from a few reference glyphs.
Azadi et al. proposed MC-GAN \cite{azadi2018multi}, which generates all 26 letters of the Latin alphabet with color and texture from five samples.
However, it is structurally difficult to create fonts for languages with a large number of characters through this method.
SA-VAE proposed by Sun et al. \cite{Sun2018Learning} is a Chinese-specific model leveraging domain knowledge, such as structures and radicals.
Zhang et al. proposed EMD \cite{zhang2018separating}, which synthesizes Chinese black-and-white and shape-distinctive fonts. Moreover, this method is applicable to other language fonts or domains.
AGIS-Net proposed by Gao et al. \cite{Gao2019Artistic} generates colored and textured Latin and Chinese fonts.
This method was developed based on EMD with adversarial training schemes.
The authors also attempted to generate black-and-white fonts with promising results, but there is room for improvement.
DM-Font proposed by Cha et al. \cite{cha2020few} utilizes dual memory banks for languages whose characters can be decomposed into a fixed number of sub-glyphs, such as Korean and Thai.
However, this system requires additional knowledge about the compositions of each glyph, which may be expensive.
Furthermore, its applicability to scripts that are not fully compositional such as Chinese has not yet been verified.


\subsection{Deep Metric Learning}
DML is a fundamental technique widely used in various areas, such as facial recognition, anomaly detection, and image retrieval.
It involves forming a feature space in which the distance between embeddings corresponds to the similarity of inputs.

Previous approaches extracted pairs (contrastive Loss by Hadsell et al. \cite{hadsell2006dimensionality}) or triplets (triplet Loss by Hoffer et al. \cite{hoffer2015deep}) from a mini-batch to move the embeddings closer to or further away from each other, focusing on their relationships.
These methods usually depend on the ability to pick a set of informative samples from a mini-batch.
As mini-batches do not necessarily reflect the distribution of the entire training data, the optimization targets are not constant, and training tends to be unstable.

Ranjan et al. proposed $L_2$-constrained softmax loss \cite{ranjan2017l2}.
They fed an $L_2$-normalized feature vector into an additional fully connected layer, whose output dimension was the number of classes.
Subsequently, they applied a softmax function and computed the cross-entropy loss.
Qian et al. \cite{qian2019softtriple} theoretically proved that minimizing a special case of the normalized softmax loss is equivalent to optimizing a smoothed triplet loss.

\section{Methods}
In this section, we introduce DML to the common component of state-of-the-art font generation frameworks: a style feature encoder.
With DML, the style encoders are expected to output embeddings that are close to each other for input glyphs having the same style (small intra-class variance), and embeddings that are far from each other for inputs with different styles (large inter-class variance).
Hence, the style feature encoders should be able to extract only style features while reducing the impact of the variances in the content features. 
Our framework is a simple but powerful approach to improve the performance of the existing schemes.

We describe the details of the loss function and the backbone frameworks below.

\subsection{Deep Metric Learning Loss}
We selected the $L_2$-constrained softmax loss proposed by Ranjan et al. \cite{ranjan2017l2} as the DML method because of its robustness and learning stability.
Let $\x_i$ be the $L_2$-normalized style feature embedding for the $i$-th sample.
$c_i$ is the font class label corresponding to $\x_i$.
$[\w_1,\ldots,\w_C]\in\R^{d\times C}$ denotes the weights of the additional fully connected layer, where $d$ is the dimension of the feature embedding, and $C$ is the number of font classes in training.
$[b_1,\ldots,b_C]\in\R^C$ represents the biases of the fully connected layer.
$\tau$ is the temperature that amplifies the difference among classes.
Consequently, we can express the loss function as follows:
\begin{equation}
  \mathcal{L}_{dml}(\x_i) = -\log\frac{\exp(( \w_{c_i}^\top \x_i + b_{c_i}) / \tau )}{\sum_j \exp(( \w_j^\top \x_i + b_j) / \tau )} .
  \label{eq:softmaxnorm}
\end{equation}

We computed the above loss function for the style feature embeddings from the models described below.

\begin{figure*}[!t]
  \centering
  \subfloat[AGIS-Net + DML]{\includegraphics[width=0.48\hsize]{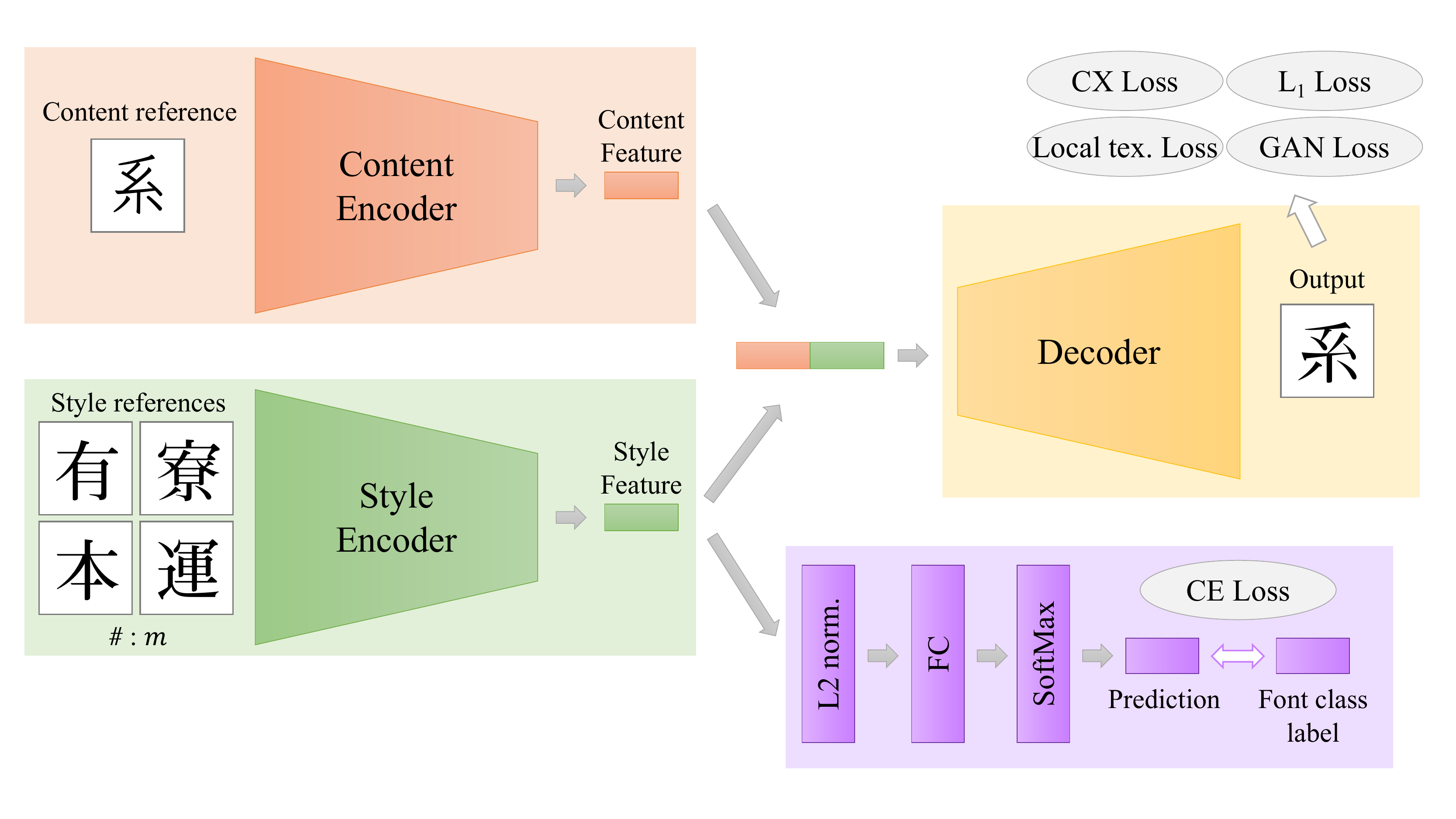}
  \label{fig:agis}}
  \hfil
  \subfloat[EMD + DML]{\includegraphics[width=0.48\hsize]{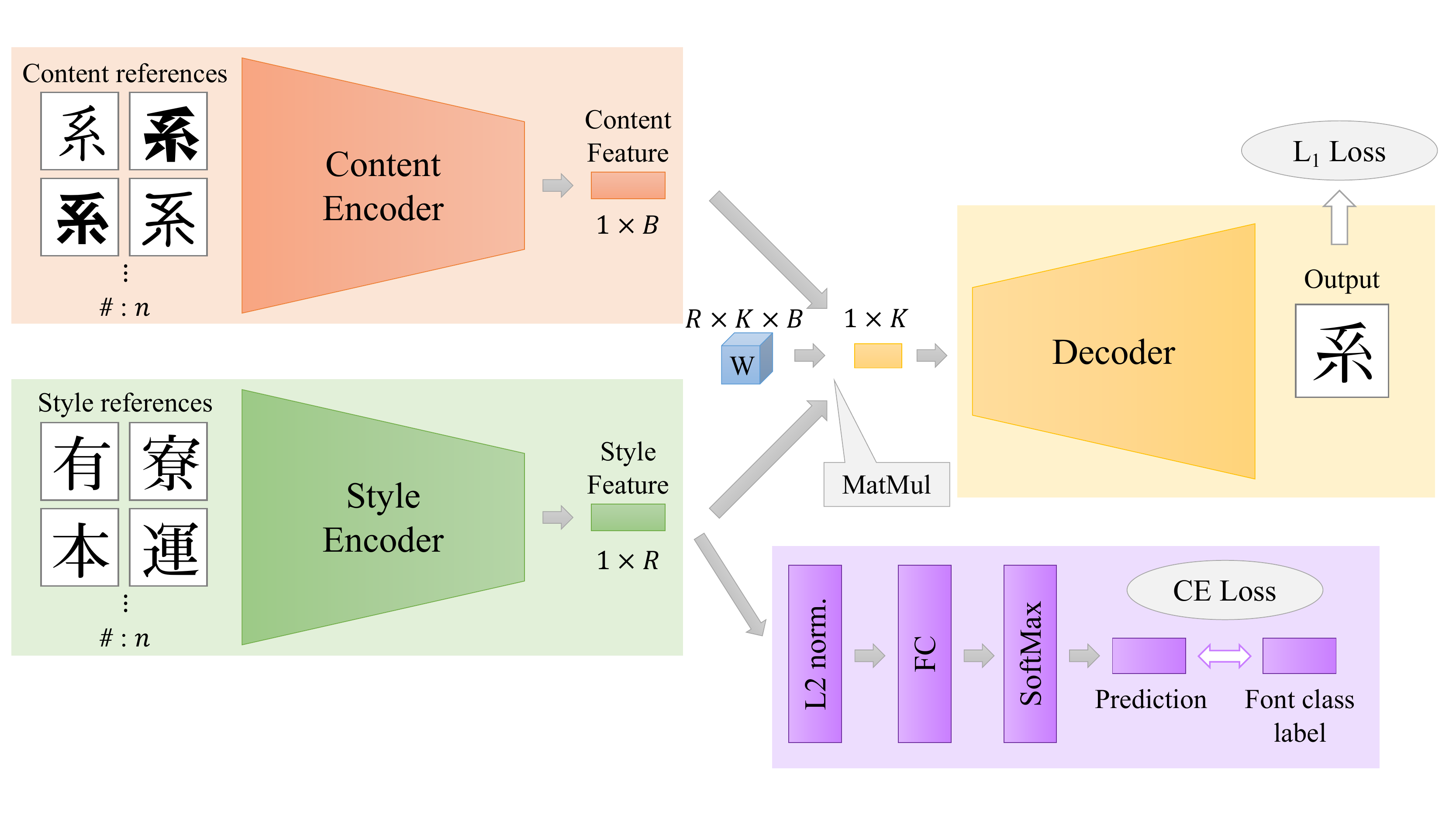}
  \label{fig:emd}}
  \caption{Generator architectures with our framework. The purple area is the branch of DML for learning better style features. Skip connections are omitted in these figures. }
  \label{fig:generator}
\end{figure*}

\subsection{Backbone Models}
\subsubsection{AGIS-Net}
\label{sec:method-agis}
AGIS-Net \cite{Gao2019Artistic} is a generic model that synthesizes colored and textured fonts.
This model follows typical GAN schemes; it has a generator and discriminators.
The generator $G$ (Fig. \ref{fig:agis}) has two CNN-based encoders: ``content encoder,'' which encodes what character to synthesize from a single content reference glyph, and ``style encoder,'' which extracts style features from a subset ($m$ glyphs) of style reference glyphs.
The font of the content reference glyphs is constant for all the training and inference processes.
For each iteration, $m$ randomly selected glyphs from among the entire style reference set ($n$ glyphs) are fed into the style encoder.
The decoders originally have two branches: ``shape decoder'' and ``texture decoder.''
As we generated black-and-white fonts in our experiments, calculation of the loss function for the intermediate black-and-white images in the shape decoder is not required.
Therefore, we eliminated the shape decoder and employ the texture decoder.

For the discriminators, the original model has a shape discriminator $D_{sha}$, texture discriminator $D_{tex}$, and local discriminator $D_{local}$.
The shape and texture discriminators attempt to distinguish whether the outputs of the shape and texture decoders, respectively, are real.
As the shape decoder is eliminated, we cannot use the shape discriminator in our training.
However, the generated glyph images are not degraded because the shape discriminator functions as a discriminator for both shapes and textures.
Regarding the local discriminator, Gao et al. proposed to cut patches randomly from glyph images and feed them to the local discriminator.
Furthermore, to obtain better texture details, they manually blurred some real samples with a Gaussian filter and regarded them as fake samples when training.

In AGIS-Net, four types of loss functions were used, in addition to our metric learning loss: $L_1$ loss, adversarial loss, contextual loss, and local texture refinement loss.
The $L_1$ loss is the pixel-wise distance between the generated images $y$ and the ground-truth images $\hat{y}$, which is defined as
\begin{equation}
  \mathcal{L}_1 = \E_{y, \hat{y}} ||y-\hat{y}||_1 .
  \label{eq:l1loss}
\end{equation}
The adversarial loss is defined as
\begin{equation}
  \mathcal{L}_{adv} = \E_y [\log(1-D_{tex}(y))] ,
  \label{eq:advloss}
\end{equation}
where $D_{tex}$ denotes the texture discriminator.
The loss of the texture discriminator is defined as
\begin{equation}
  \mathcal{L}(D_{tex}) = \E_{t^r_s} [\log(D_{tex}(t^r_s))] + \E_y [\log(1-D_{tex}(y))] ,
  \label{eq:advDloss}
\end{equation}
where $t^r_s$ is a real glyph image.
The contextual loss \cite{mechrez2018contextual} was used to measure the similarity between two images, without spatial alignment.
Assume that $CX(Y, \hat{Y})$ is the similarity between two feature maps $Y$ and $\hat{Y}$, $\Phi^l(\cdot)$ is the $l$-th layer feature of VGG19 \cite{simonyan2014very}, and $L$ is the number of used layers.
The contextual loss is defined as
\begin{equation}
  \mathcal{L}_{CX} = -\frac{1}{L} \sum_l \log (CX (\Phi^l(y), \Phi^l(\hat{y}))) .
  \label{eq:cxloss}
\end{equation}
The local texture refinement loss is defined as
\begin{equation}
  \mathcal{L}_{local} = \E_{p_y} [\log (1 - D_{local}(p_y) ) ] ,
  \label{eq:localloss}
\end{equation}
\begin{equation}
  \begin{split}
    \mathcal{L}(D_{local}) = \ &\E_{p_{real}} [\log(D_{local}(p_{real}))] + \\ & \E_{p_{blur}} [\log(1 - D_{local}(p_{blur}))] + \\ & \E_{p_y} [\log (1 - D_{local}(p_y) ) ] ,
  \end{split}
  \label{eq:localDloss}
\end{equation}
where $p_y$ and $p_{real}$ are the patches from the generated images and style reference images, respectively, and $p_{blur}$ indicates the blurred patches of $p_{real}$.

We used these loss functions as described in the original paper, except for the shape-decoder-related losses.
Thus, the loss function used to optimize the generator is expressed as
\begin{equation}
  \begin{split}
    \mathcal{L}(G) = \ &\lambda_{L_1} \mathcal{L}_1 + \lambda_{adv} \mathcal{L}_{adv} + \lambda_{CX} \mathcal{L}_{CX} + \\ &\lambda_{local} \mathcal{L}_{local} + \lambda_{dml} \mathcal{L}_{dml} ,
  \end{split}
  \label{eq:Gloss}
\end{equation}
where $\lambda_{\bullet}$ denotes the weights used for balancing the losses.
The final objective used to optimize the generator $G$ and the discriminators $D_{tex}, D_{local}$ is
\begin{equation}
  \begin{split}
    \min_G \max_{D_{tex}, D_{local}} \mathcal{L}(G, D_{tex}, D_{local}).
  \end{split}
  \label{eq:agis-obj}
\end{equation}

The training process has two phases: pre-training and fine-tuning.
In the pre-training phase, we performed supervised and adversarial learning using several fonts.
In the subsequent fine-tuning phase, we performed few-shot training for a single unseen font that we intended to generate.
Note that only $n$ style reference glyph images are available in this stage.
If the glyph to be generated is \textit{not} in the few-shot reference set, we cannot view its ground-truth image.
Then, $\mathcal{L}_1$ and $\mathcal{L}_{CX}$ are set to zero because they are pair-wise loss functions that require ground-truth images.
Furthermore, we did not use the metric learning loss $\mathcal{L}_{dml}$ when fine-tuning the model.

\subsubsection{EMD}
EMD \cite{zhang2018separating} is also a generic model to synthesize glyphs, but it does not employ the GAN architecture.
The generator architecture of EMD (Fig. \ref{fig:emd}) is similar to that of AGIS-Net, but it utilizes the bilinear model as the feature mixer.
The input format is also different, and EMD concatenates all the $n$ images of the content/style reference set to feed them to each encoder.
The training process is one-stage supervised learning, in which we used the entire training dataset.
Then, in the inference stage, we generated glyphs from $n$ content reference glyphs within the training dataset and $n$ style reference glyphs having novel styles.

The loss function for the supervised training is expressed as follows:
\begin{equation*}
  \alpha^{\hat{y}} = 1 / N_b^{\hat{y}} ,
\end{equation*}
\begin{equation*}
  \beta^{\hat{y}} = \frac{\exp({\rm mean}_{\hat{y}})}{\sum_{\hat{y} \in \mathcal{D}_t} \exp({\rm mean}_{\hat{y}})} ,
\end{equation*}
\begin{equation}
  \mathcal{L}_1 = \alpha^{\hat{y}} \beta^{\hat{y}} ||y - \hat{y}||_1 ,
  \label{eq:emdl1loss}
\end{equation}
where $N_b^y$ is the number of black pixels in $y$, $\mathcal{D}_t$ denotes the training dataset, and ${\rm mean}_{i}$ is the mean value of the black pixels in the image $i$.

The final objective used to optimize the generator $G$ is
\begin{equation}
  \begin{split}
    \min_G \mathcal{L}_1 + \lambda_{dml} \mathcal{L}_{dml}.
  \end{split}
  \label{eq:emd-obj}
\end{equation}

\section{Experiment}

\subsection{Dataset}
We constructed a Japanese typographic font dataset consisting of 368 fonts and 2965 glyphs to evaluate the effectiveness of our proposed framework.
All the images are grayscale and have the dimensions 64 $\times$ 64 pixels.
We randomly selected 338 fonts for the pre-training of AGIS-Net and the training of EMD.
Among these 338 fonts, we used 165 glyphs for the validation set and 2800 glyphs for training.
Furthermore, we used 30 fonts not used in training for the fine-tuning of AGIS-Net and the inference of AGIS-Net/EMD.
For each font, $n \in \{5, 10, 15, 30\}$ style reference glyphs are available when synthesizing the unseen style font.

\begin{figure}[!t]
  \centering
  \includegraphics[width=2.8in]{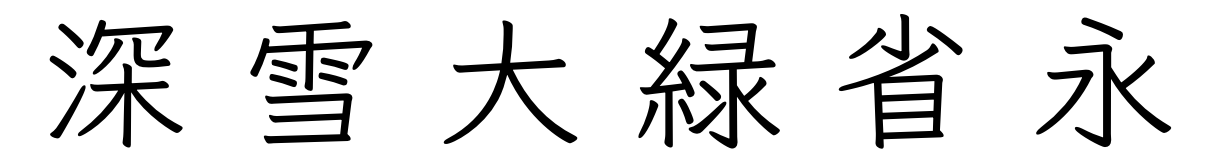}
  \caption{Sample images of the font used as the content reference for AGIS-Net}
  \label{fig:content_sample}
\end{figure}

As mentioned in section \ref{sec:method-agis}, the font of the content reference glyphs is constant for all the training and inference processes of AGIS-Net.
We selected a standard and simple style font, whose examples are shown in Fig. \ref{fig:content_sample}.
Note that we did not include this font in the 368 fonts.

\subsection{Implementation Details}
\label{sec:impl}

The weight of the metric learning loss $\lambda_{dml}$ was set to 1.0.
The temperature of softmax $\tau$ was selected as 0.5 for the pre-training of AGIS-Net and 0.1 for the training of EMD.
The other hyperparameters and the settings of the convolutional layers are the same as the original settings.

We performed the pre-training of AGIS-Net for 20 epochs.
The learning rate in the first 10 epochs was set to 0.0002 and was reduced linearly in the remaining 10 epochs.
Subsequently, we performed fine-tuning for 200 epochs at the learning rate of 0.00002.
The number of glyphs fed to the style encoder simultaneously, $m$, is set to 4.
Note that the dimensionalities of the content/style feature embeddings are 512.

Furthermore, we performed the training of EMD for 10 epochs at the learning rate of 0.0002.
The dimensionalities of the feature embeddings, $R$, $B$, and $K$ (Fig. \ref{fig:emd}) are all 512.

\subsection{Results}

\begin{figure*}[!t]
  \centering
  \includegraphics[width=\hsize]{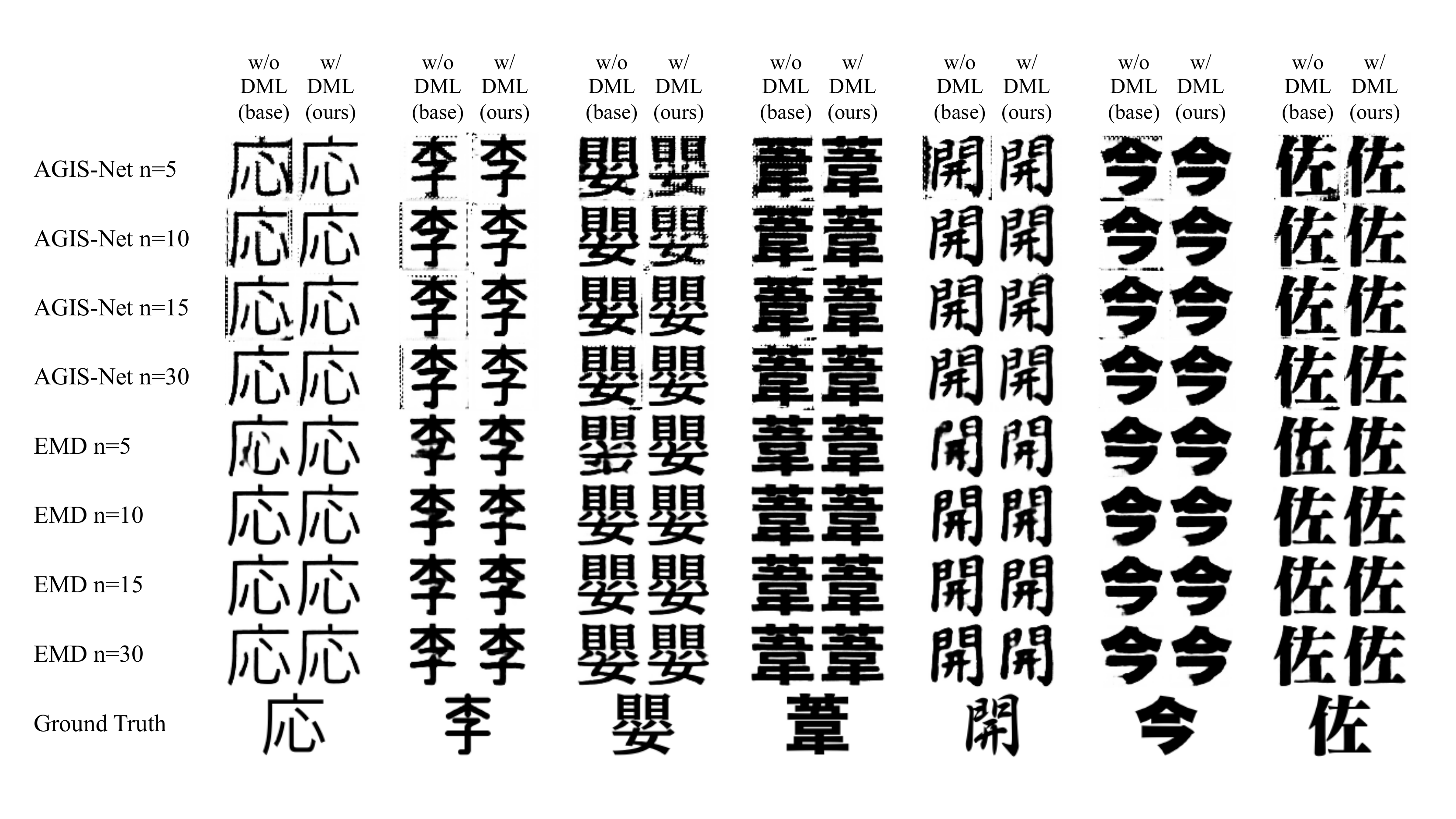}
  \caption{Visual comparison of AGIS-Net \cite{Gao2019Artistic} and EMD \cite{zhang2018separating} with and without DML}
  \ \\
  \label{fig:result}
\end{figure*}

\begin{table*}[!t]
  \renewcommand{\arraystretch}{1.3}
  \setlength{\tabcolsep}{6pt}
  \caption{Quantitative Comparison of AGIS-Net \cite{Gao2019Artistic} and EMD \cite{zhang2018separating} with and without DML}
  \label{table:quant}
  \centering
  \begin{tabular}{lccccccccc}
    \hline
    \multirow{2}{*}{Method}& \multicolumn{4}{c}{$n=5$} && \multicolumn{4}{c}{$n=10$} \\
    & $L_1$ Loss & PSNR & SSIM & FID && $L_1$ Loss & PSNR & SSIM & FID \\
    \hline
    AGIS-Net \cite{Gao2019Artistic}       & 0.306 & 9.85 & 0.554 & 123.82 && 0.275 & 10.53 & 0.607 & 89.33 \\
    AGIS-Net \cite{Gao2019Artistic} + DML (ours) & \textbf{0.274} & \textbf{10.44} & \textbf{0.566} & \textbf{115.41} && \textbf{0.219} & \textbf{11.84} & \textbf{0.656} & \textbf{69.76} \\
    \hline
    EMD \cite{zhang2018separating}        & 0.189 & 13.48 & 0.701 & 63.79 && \underline{\textbf{0.179}} & 13.87 & 0.721 & 57.84 \\
    EMD \cite{zhang2018separating} + DML (ours)  & \underline{\textbf{0.187}} & \underline{\textbf{13.58}} & \underline{\textbf{0.705}} & \underline{\textbf{60.91}} && \underline{\textbf{0.179}} & \underline{\textbf{13.96}} & \underline{\textbf{0.726}} & \underline{\textbf{54.59}} \\
    \hline\\
  \end{tabular} 
  \begin{tabular}{lccccccccc}
    \hline
    \multirow{2}{*}{Method}& \multicolumn{4}{c}{$n=15$} && \multicolumn{4}{c}{$n=30$} \\
    & $L_1$ Loss & PSNR & SSIM & FID && $L_1$ Loss & PSNR & SSIM & FID \\
    \hline
    AGIS-Net \cite{Gao2019Artistic}       & 0.245 & 11.07 & 0.640 & 77.78 && 0.208 & 12.29 & 0.684 & 60.14 \\
    AGIS-Net \cite{Gao2019Artistic} + DML (ours) & \textbf{0.202} & \textbf{12.56} & \textbf{0.686} & \underline{\textbf{53.72}} && \textbf{0.193} & \textbf{13.03} & \textbf{0.705} & \underline{\textbf{46.86}} \\
    \hline
    EMD \cite{zhang2018separating}        & 0.176 & 13.95 & 0.727 & 57.81 && 0.174 & 14.04 & 0.733 & \textbf{53.82} \\
    EMD \cite{zhang2018separating} + DML (ours)  & \underline{\textbf{0.174}} & \underline{\textbf{14.07}} & \underline{\textbf{0.730}} & \textbf{57.41} && \underline{\textbf{0.170}} & \underline{\textbf{14.25}} & \underline{\textbf{0.738}} & 53.84 \\
    \hline
  \end{tabular}
\end{table*}

We synthesized Japanese typographic fonts from $n \in \{5, 10, 15, 30\}$ style reference glyphs with AGIS-Net and EMD.
The generated glyphs are shown in Fig. \ref{fig:result}.
The results demonstrated that our method produces glyphs with clearer contours and less noise compared with the baseline methods.
The baseline methods generated poor results, especially when $n$ is extremely small, whereas the enhanced models with the proposed method produced accurate results.

Moreover, we conducted quantitative evaluations of the proposed method.
To evaluate the performance, we adopted four commonly used metrics: $L_1$ loss, peak signal-to-noise ratio (PSNR), structural similarity index (SSIM) \cite{wang2004image}, and Frechet inception distance (FID) \cite{heusel2017gans}.
The smaller the values of $L_1$ loss and FID, the better is the performance.
In contrast, the larger the values of PSNR and SSIM, the better is the performance.
We calculated these metrics for each of the 30 fonts and then calculated the average of these values, as summarized in Table \ref{table:quant}.
The methods enhanced with our framework outperformed the baseline methods in almost all the metrics and $n$.

Our framework with DML is more effective when the number of style reference glyphs is small, which leads to the following inferences.
Introducing DML allows our model to extract only the font style features, without being affected by the differences in contents.
Therefore, our method requires fewer style reference glyphs for capturing the style characteristics of novel fonts.

\subsection{Effects of Deep Metric Learning}

\begin{table}[!t]
  \renewcommand{\arraystretch}{1.3}
  \caption{Quantitative Evaluation of Style Feature Separation}
  \label{table:feature}
  \centering
  \begin{tabular}{lccc}
    \hline
    Method & R@1 & R@2 & NMI \\
    \hline
    AGIS-Net \cite{Gao2019Artistic}              & 0.7556 & 0.8689 & 0.8510 \\
    AGIS-Net \cite{Gao2019Artistic} + DML (ours) & \textbf{0.7822} & \textbf{0.8756} & \textbf{0.8628} \\
    \hline
  \end{tabular}
\end{table}

We investigated the distribution of style feature embeddings using AGIS-Net to verify whether the style encoder can output better style feature embeddings with our framework.
We pre-trained AGIS-Net with and without DML by using the settings mentioned in Section \ref{sec:impl}.
Then, we extracted style feature embeddings from 30 style reference glyphs for each of 30 unseen fonts, with the pre-trained style encoder.
As the AGIS-Net style encoder architecture requires $m\ (=4)$ images to be input simultaneously, we concatenated a single image $m$ times.

With the obtained 900 style feature embeddings, we evaluated the following two metrics: recall@$k$ \cite{jegou2010product} and normalized mutual information (NMI) \cite{NMI}.
Recall@$k$, a standard measure in image retrieval, denotes the proportion of query vectors for which the embedding vectors of the same class appear in the nearest $k$ neighbors in Euclidean metrics.
For NMI, we clustered the style feature embeddings into 30 classes with k-means++ \cite{arthur2007kmeans} for 100 times, calculated the NMI between each result of the clustering and the true classes, and reported the best value.
The results are listed in Table \ref{table:feature}, where R@$k$ indicates Recall@$k$.
All the metrics indicate that our framework with DML helps in improving feature separation.

\begin{figure*}[!t]
  \centering
  \subfloat[AGIS-Net with DML]{\includegraphics[width=\hsize]{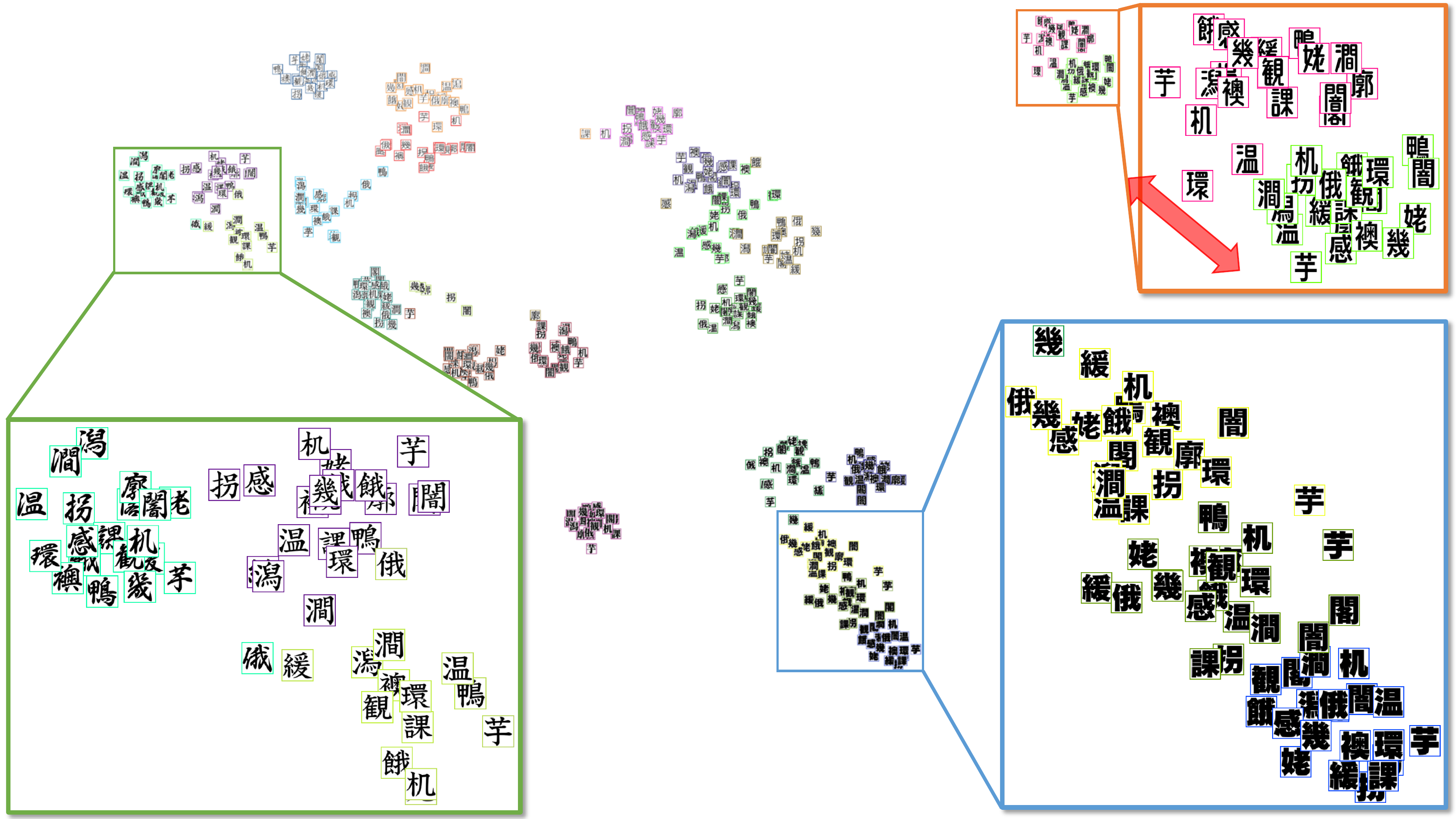}
  \label{fig:tsne_agis}}
  \\
  \bigskip
  \subfloat[AGIS-Net without DML]{\includegraphics[width=\hsize]{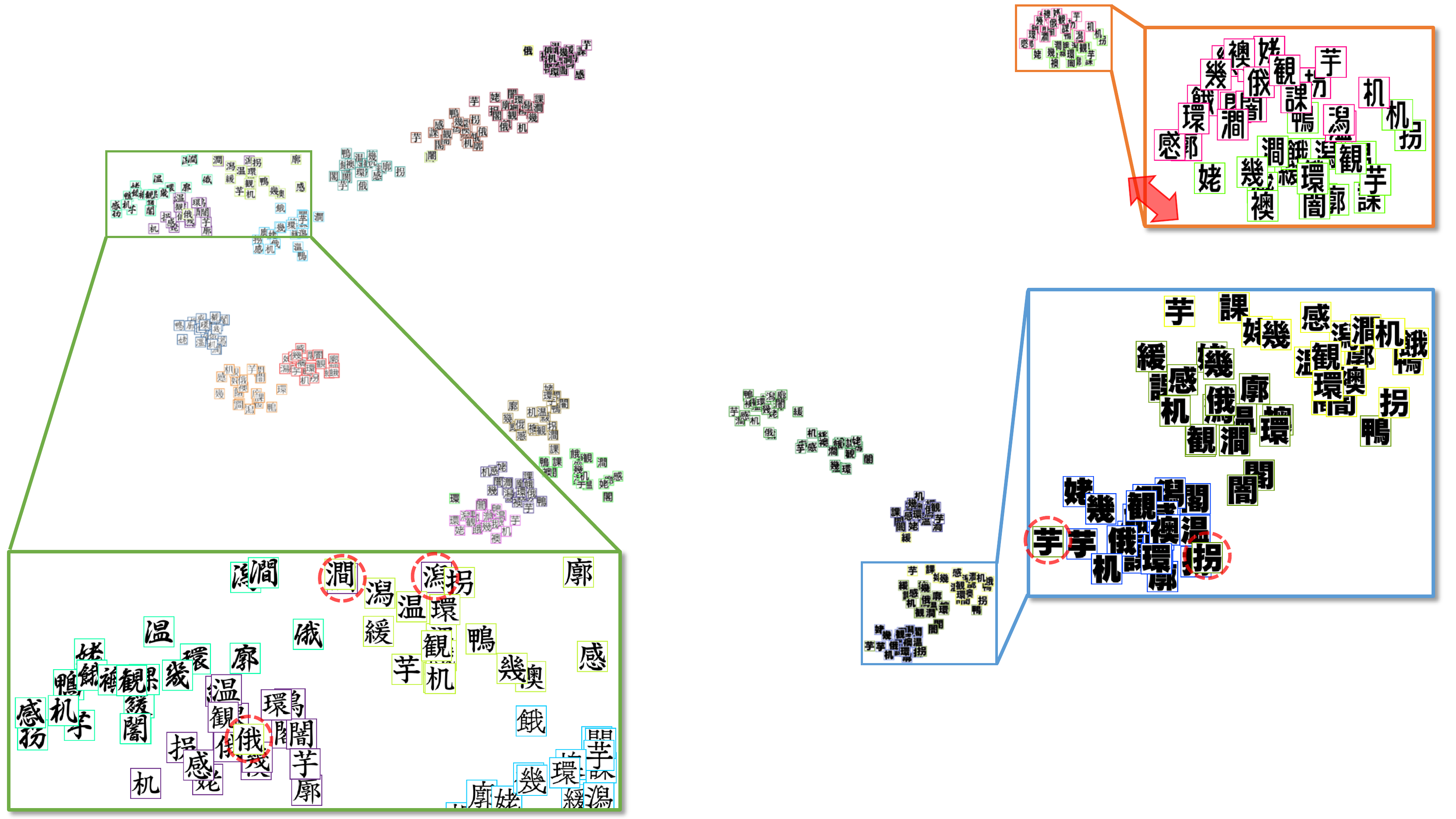}
  \label{fig:tsne_base}}
  \caption{t-SNE projection of style feature embeddings. We visualized a portion of all the glyphs for simplicity. Different colors denote different font styles. Comparing the results between with and without DML, we can observe the distances between different fonts are larger in DML, as shown by red double-sided arrows. Dashed circles show wrong clustering in the result without DML.}
  \label{fig:tsne}
\end{figure*}

Moreover, we visualized these style feature embeddings with t-distributed stochastic neighbor embedding (t-SNE) \cite{maaten2008visualizing} (Fig. \ref{fig:tsne}).
With the use of DML, different styles exist far away from each other, whereas similar styles are located close together but with less overlap.
This suggests that the style encoder can extract features by focusing on the differences in style, even in cases where the differences in style are slighter than the differences in content.
Such improved feature separation leads to an improvement in the results of typographic font generation.

\section{Conclusion}

In this study, we proposed a powerful and simple framework to improve the performance of existing few-shot font generation methods, handling Japanese typographic fonts.
The existing methods are known to produce poor results when the number of style reference glyphs is extremely limited.
We introduced DML through $L_2$-constrained softmax loss to the style encoder and demonstrated the remarkable improvement in the outcomes.
We used AGIS-Net and EMD as the baseline methods in this study.
However, our framework is general and is easily applicable to other methods and tasks.
Furthermore, experimental results indicated that our framework helps style encoders focus on style characteristics, without being biased by content properties.
Consequently, promising results can be obtained when the style reference glyphs are considerably limited.






\bibliographystyle{IEEEtran}
\bibliography{IEEEabrv,ref}
%



\end{document}